%% file: 00paper.tex
\documentclass[sigconf,nonacm,natbib=true]{acmart}
\usepackage{graphicx}
\usepackage{subcaption}
\AtBeginDocument{%
  \providecommand\BibTeX{{%
    \normalfont B\kern-0.5em{\scshape i\kern-0.25em b}\kern-0.8em\TeX}}}

\setcopyright{cc}
\copyrightyear{2025}
\acmYear{2025}
\acmDOI{XXXXXXX.XXXXXXX}

\usepackage{graphicx}
\usepackage{booktabs} 
\usepackage{color}
\usepackage{multirow}
\usepackage{pifont}
\usepackage{graphicx}

\newcommand{\cmark}{\ding{51}}

\newcommand{\goodarrow}{\raisebox{-0.2ex}{\scalebox{0.6}{\textcolor{green!60!black}{\ensuremath{\uparrow}}}}}
\newcommand{\badarrow}{\raisebox{-0.2ex}{\scalebox{0.6}{\textcolor{red!70!black}{\ensuremath{\downarrow}}}}}
\newcommand{\nochange}{\phantom{\raisebox{-0.2ex}{\scalebox{0.6}{\ensuremath{\uparrow}}}}}
\newcommand{\dataset}{MultiConAD}

\newcommand{\captionshrink}{\vspace*{-0.5\baselineskip}}

\usepackage{enumitem}
\setlist[itemize]{leftmargin=*}
\begin{document}

\fancyhead{}
\title{MultiConAD: A Unified Multilingual Conversational Dataset for Early Alzheimer's Detection} 

\author{Arezo Shakeri}
\affiliation{%
  \institution{University of Stavanger}
  \city{Stavanger}
  \country{Norway}
}
\email{arezo.shakeri@uis.no}

\author{Mina Farmanbar}
\affiliation{%
  \institution{University of Stavanger}
  \city{Stavanger}
  \country{Norway}
}
\email{mina.farmanbar@uis.no}

\author{Krisztian Balog}
\affiliation{%
  \institution{University of Stavanger}
  \city{Stavanger}
  \country{Norway}
}
\email{krisztian.balog@uis.no}

\begin{abstract}

Dementia is a progressive cognitive syndrome with Alzheimer’s disease (AD) as the leading cause.
Conversation-based AD detection offers a cost-effective alternative to clinical methods, as language dysfunction is an early biomarker of AD. However, most prior research has framed AD detection as a binary classification problem, limiting the ability to identify Mild Cognitive Impairment (MCI)—a crucial stage for early intervention. Also, studies primarily rely on single-language datasets, mainly in English, restricting cross-language generalizability. To address this gap, we make three key contributions. First, we introduce a novel, multilingual dataset for AD detection by unifying 16 publicly available dementia-related conversational datasets. This corpus spans English, Spanish, Chinese, and Greek, and incorporates both audio and text data derived from a variety of cognitive assessment tasks. Second, we perform finer-grained classification, including MCI, and evaluate various classifiers using sparse and dense text representations. Third, we conduct experiments in monolingual and multilingual settings, finding that some languages benefit from multilingual training while others perform better independently. This study highlights the challenges in multilingual AD detection and enables future research on both language-specific approaches and techniques aimed at improving model generalization and robustness.

\end{abstract}


\maketitle

\input{dementia-01}

\input{dementia-02}
\input{dementia-03}

\input{dementia-04}

\input{dementia-05}

\bibliographystyle{ACM-Reference-Format}
\bibliography{references.bib}

\end{document}

%% file: dementia-01.tex
\section{Introduction}
  
Dementia is a progressive clinical cognitive decline that impairs daily activities and independence~\cite{grasset2018evolution}. Cases are expected to rise from 57 million in 2019 to 152 million by 2050~\cite{guo2024trajectory}. According to the World Health Organization, in 2019, it was the seventh leading cause of death globally and the second in high-income countries~\cite{johnsen2023incidence}. Alzheimer’s disease (AD) accounts for 50\%–75\% of cases with prevalence roughly doubling every five years after the age 65~\cite{lane2018alzheimer}. The lack of effective medical treatments underscores the need for early dementia and prevention~\cite{guo2024trajectory}.

Traditional clinical methods for AD detection, including MRI, PET imaging, and cerebrospinal fluid analysis, are costly, time-consuming, and impractical for large-scale early screening. This has led to efforts to find more accessible and cost-effective alternatives~\cite{yang2022deep}. Recent studies have increasingly identified language dysfunction as an early indicator of cognitive decline, making speech and language features as valuable biomarkers for the early detection of dementia~\cite{petti2020systematic, Language2024}. This growing body of research suggests that analyzing conversational patterns can provide critical insights into the onset of AD and other forms of dementia. However, a major limitation of existing studies is their reliance on single-language datasets, predominantly in English~\cite{perez2022alzheimer, NCMMSC2021, Agbavor2022plos}, with frequently utilized datasets including DementiaBank’s Pitt corpus, ADReSS, and ADReSSo.  This focus on only one language restricts the cross-linguistic generalizability of findings, making it difficult to apply these methods in multilingual or culturally diverse settings. Furthermore, the lack of multilingual datasets poses challenges for developing diagnostic tools that can be used globally~\cite{fraseretal2019multilingual}.
Expanding this scope, the recent Interspeech TAUKADIAL Challenge aims to investigate speech as a marker of cognitive function in a global health context~\cite{Benjamin2024}. This initiative provides data in two major languages, Chinese and English, to enhance understanding of the relationship between speech and cognition across diverse linguistic settings.
The Signal Processing Grand Challenge (SPGC) is another effort aimed at exploring the transferability of speech features across languages (English and Greek) for AD prediction~\cite{luz2023multilingual}. %
Despite these efforts, cross-linguistic speech-based detection of AD remains an open gap in the field. Additionally, most prior research has framed AD detection as a binary classification problem (AD vs. Healthy Controls (HC)), limiting the ability to identify Mild Cognitive Impairment (MCI)---a crucial intermediate stage for early intervention~\cite{shakeri2025natural}. Detecting MCI is particularly important because individuals at this stage have a higher likelihood of progressing to AD, yet timely interventions can potentially slow or prevent further decline~\cite{Knopman2014mayo}. However, the lack of datasets with labeled MCI cases has restricted the development of models that can distinguish between these three cognitive states.

To address this significant gap, we present MultiConAD, a unified, multilingual conversational dataset for Alzheimer's detection. MultiConAD consolidates 16 publicly available dementia-related conversational datasets across four languages: English, Spanish, Chinese, and Greek. These languages represent some of the most widely spoken languages in the world, with Mandarin Chinese and Spanish being the most spoken, followed by English~\cite{shakeri2025natural}. The corpus incorporates both audio and text modalities; most datasets include both, while a few contain only audio or only text. Crucially, these datasets feature a diverse range of cognitive assessment tasks, such as picture descriptions, story recall tasks, and verbal and semantic fluency tests.
As our second contribution, we perform a more nuanced classification approach at a finer level of granularity by considering MCI as an intermediate category between AD and HC. Third, we conduct extensive experiments on variants of the dataset (monolingual, multilingual, and translated) using different text representations and classification algorithms. The overarching objective of this research is to investigate the impact of classification approach (binary vs. multiclass), language, and dataset composition (monolingual, multilingual, translated) on the performance of Alzheimer's detection models using conversational data.

Our results reveal interesting language-specific trends: while some languages benefit from multilingual training, suggesting shared markers of cognitive decline, other languages perform better when trained separately, indicating unique language-dependent patterns. 
These findings highlight the importance of tailoring AD detection models to specific language while also demonstrating the potential advantages of leveraging multilingual training for improved generalization. 
Beyond these immediate findings, MultiConAD faciliates future research in several key areas, including optimizing language-specific models through deeper linguistic analysis and leveraging cross-lingual transfer learning.

In summary, this paper presents the following contributions: (1) we introduce and release a unified multilingual conversational dataset for early Alzheimer’s detection; (2) we experimentally evaluate and compare various approaches for leveraging monolingual, multilingual, and translated datasets for both binary and multiclass classification tasks; and (3) we offer a thorough analysis of the results, emphasizing the key challenges associated with this problem. All resources developed in this study, including instructions on how to obtain the dataset and model implementations, are publicly accessible at \url{https://github.com/ArezoShakeri/MultiConAD}.

%% file: dementia-02.tex
\section{Related work}
\label{sec:related}

To provide the necessary context for the unified Alzheimer's detection dataset introduced in this paper, we discuss cognitive tasks used for Alzheimer's detection, datasets, and computational approaches.

\subsection{Cognitive Task Analysis in Alzheimer's Detection}
    
Cognitive Task Analysis is an essential method used in AD studies to evaluate various aspects of communication, memory, executive function, and overall cognitive ability as the disease progresses \cite{Guarino2019Aging}. Boston Diagnostic Aphasia Examination (BDAE) is a well-known method for assessing linguistic abilities and language impairments through tasks such as word fluency tests, sentence construction, and narrative storytelling, helping clinicians and researchers understand how language deficits manifest in different stages of Alzheimer's \cite{Goodglass:1972:Book}. One widely used component of the BDAE is the \textbf{Cookie Theft Picture} (CTP) \cite{Cummings2019Describing}, an image depicting a busy kitchen scene where participants are asked to describe the situation in detail. This task assesses a person's ability to produce coherent speech and reveals cognitive and language dysfunctions, such as difficulties in organizing thoughts or recalling key details from memory.
Another prominent task is \textbf{Story Recall}, where participants are instructed to memorize the short story to be told, and are asked to recall the story when it ends \cite{Wechsler:2009:Book}. Additionally, \textbf{Verbal Fluency Tests} are commonly used to evaluate an individual’s lexical retrieval and executive function \cite{lezak2004neuropsychological}. These tasks require participants to generate as many words as possible within a time limit, either from a specific letter (e.g., F, A, S) or a semantic category (e.g., animals, fruits). The \textbf{Semantic Fluency Task} is also widely employed in clinical settings to identify challenges in speech production, executive functioning, and semantic memory performance \cite{lezak2004neuropsychological}. In this task, participants are asked to produce as many words as possible in a given semantic category (e.g., animals) and time frame.

\begin{table*}[ht]
  \centering
  \caption{Overview of datasets used in this study. Tasks: (PD) Picture Description, (FT) Fluency Task, (SR) Story Retelling, (FC) Free Conversation, (NA) Narrative. Labels: (DM) Dementia, (AD) Alzheimer’s Disease, (MCI) Mild Cognitive Impairment, (HC) Healthy Control.  
   }
   \captionshrink
    \begin{tabular}{cl|cc|ccccc|cccc}
    \toprule
    \multirow{2}{*}{\textbf{Language}} & \multirow{2}{*}{\textbf{Source}} & \multicolumn{2}{c|}{\textbf{Modality}} & \multicolumn{5}{c|}{\textbf{Task}} & \multicolumn{4}{c}{\textbf{Labels}}    \\
      &       & Text  & Audio & PD    & FT    & SR    & FC    & NA    & DM    & AD    & MCI   & HC    \\
    \midrule
    \multirow{8}[2]{*}{English} & Pitt  & \cmark      & \cmark     & \cmark     & \ding{55}      & \ding{55}     & \ding{55}     & \ding{55}     & \ding{55}     & 255   & 42    & 243      \\
          & Lu    & \cmark     & \cmark     & \cmark     & \ding{55}     & \ding{55}     & \ding{55}     & \ding{55}     & 6     & 16    & 2     & 27     \\
          & VAS   & \cmark     & \cmark     &   \ding{55}    & \ding{55}     & \ding{55}     & \cmark     & \ding{55}     & 30    & \ding{55}     & 35    & 36      \\
          & Baycrest & \cmark     & \cmark     &  \ding{55}     &  \ding{55}     &   \cmark    &   \ding{55}      &   \ding{55}    & \ding{55}     & 3     & 7     & \ding{55}       \\
          & Kempler & \cmark     & \cmark     & \cmark     & \ding{55}     & \ding{55}     & \ding{55}     & \ding{55}     & \ding{55}     & 7     & \ding{55}     & \ding{55}       \\
          & WLS   & \cmark     & \cmark     & \cmark     & \cmark     & \ding{55}     & \ding{55}     & \ding{55}     & \ding{55}    & 263    & \ding{55}     & 1106  \\
          & Delware & \cmark     & \cmark     & \cmark     & \ding{55}     & \ding{55}     & \ding{55}     & \cmark     & \ding{55}     & \ding{55}     & 61    & 34     \\
          & Taukdial & \ding{55}     & \cmark     & \cmark     & \ding{55}     & \ding{55}     & \ding{55}     & \ding{55}     & \ding{55}     & \ding{55}     & 95    & 74     \\
    \midrule
    \multirow{2}[2]{*}{Spanish} & Ivanova & \cmark     & \cmark     &  \ding{55}      &   \ding{55}    &    \ding{55}    &  \ding{55}      &   \cmark    & \ding{55}     & 74    & 90    & 197     \\
          & PerLA & \cmark     & \ding{55}     & \cmark     & \cmark     & \ding{55}     & \cmark     & \ding{55}     & \ding{55}     & 21    & \ding{55}     & \ding{55}       \\
    \midrule
    \multirow{2}[2]{*}{Chineas} & NCMMSE & \ding{55}     & \cmark     & \cmark     & \cmark     & \ding{55}     & \cmark     & \ding{55}     & \ding{55}     & 79    & 93    & 108     \\
          & iFkyTek & \cmark     & \ding{55}     & \cmark     & \ding{55}     & \ding{55}     & \ding{55}     & \ding{55}     & \ding{55}     & 68    & 144   & 111     \\
    \midrule
    \multirow{4}[2]{*}{Greek} & DS3   &   \ding{55}    &  \cmark     &    \cmark   &   \cmark    &    \ding{55}   &  \ding{55}     &      \ding{55} & \ding{55}     & 76    & \ding{55}     & 19     \\
          & DS5   &    \ding{55}   &    \cmark    &    \cmark    &   \cmark     &    \ding{55}   &   \ding{55}    &   \ding{55}    & \ding{55}     & 26    & 35    & 31      \\
          & DS7   &    \ding{55}    &   \cmark     &   \cmark     &   \cmark     &   \ding{55}     &   \ding{55}     &   \ding{55}     & \ding{55}     & 27    & 35    & 2       \\
          & ADReSS-M &   \ding{55}    &   \cmark     &    \cmark    &  \ding{55}     &\ding{55}&  \ding{55}     &    \ding{55}   & \ding{55}     & 22    & \ding{55}     & 24     \\
    \bottomrule
    \end{tabular}%
  \label{tab:datasetdescription}%
\end{table*}%

\subsection{Datasets}

The dataset developed in this study is multilingual, comprising four languages:  Chinese, Spanish, English, and Greek, sourced from multiple countries and regions. The combined multilingual dataset includes participants from North America (United States), Europe (Spain, Greece), and Asia (China). This selection of languages ensures that the dataset captures a broad spectrum of speech patterns and cognitive variations seen in different linguistic groups. 

To build this diverse dataset, data were sourced from 16 publicly available resources that provide conversational speech related to AD. This corpus incorporates both audio and text formats. All datasets utilized in this study, with the exception of two Chinese datasets, are publicly accessible through DementiaBank, representing the most prominent publicly available datasets for Alzheimer's detection.
Access to DementiaBank is restricted to members of the DementiaBank consortium and is password-protected. Established researchers and clinicians specializing in dementia may apply for membership to gain access. The two Chinese datasets are available in the following GitHub repositories: \url{https://github.com/lzy1012/Alzheimer-s-disease-datasets} and \url{https://github.com/lzl32947/NCMMSC2021_AD_Competition}. 

The participant groups included in these datasets consist of individuals with dementia (DM), AD, MCI, and HC. Among these, only the two English datasets, Lu and VAS, contain data for dementia. It is important to note that dementia encompasses various types of diseases, including AD, Parkinson's disease, vascular dementia, and others. Table~\ref{tab:dataset_transcript} presents small parts from picture description conversations in the Pitt and Taukdial datasets. The Taukdial transcripts were originally in audio format and were converted to text using automatic transcription.

\begin{table}[t]
    \caption{Excerpts from transcripts of picture descriptions from the Pitt and Taukdial datasets. ``INT'' and ``PAR'' denote interviewer and participant utterances in the Pitt corpus; Taukdial has only participant-side utterances.}
    \centering
    \begin{tabular}{|p{1cm}|p{7cm}|}  %
        \hline
        \multicolumn{1}{|c|}{\textbf{Dataset}} & \multicolumn{1}{c|}{\textbf{Transcript}} \\  
        \hline
        Pitt & INT: just tell me everything that you see happening in that picture. PAR: alright. there's \&-um a young boy that's getting a cookie jar. PAR: and it he's \&-uh in bad shape because \&-uh the thing is PAR: and in the picture the mother is washin(g) dishes and doesn't see [\dots] \\  
        \hline
        Taukdial & What I see is a young lady was riding her tricycle with her cat, and somehow the cat got up in the tree. Her dog is trying to chase the cat, can't get up the tree. Her father climbed the tree and is sitting on the limb [\dots] \\  
        \hline
    \end{tabular}\label{tab:dataset_transcript}
\end{table}

Table~\ref{tab:datasetdescription} provides an overview of all datasets, followed by a detailed description of each dataset below.

\begin{table*}[t]
  \centering
  \caption{Summary of related work in Alzheimer's detection.  
  }
  \captionshrink
  \resizebox{\textwidth}{!}{%
  \begin{tabular}{lllll}
    \toprule
    \textbf{Reference} & \textbf{Dataset} & \textbf{Categories} & \textbf{Accuracy} & \textbf{Method} \\
    \midrule
    \cite{chen2023cross} & ADRReSS-M & AD, HC & 69.6\% & Acoustic, linguistic, and para-linguistic features and SVM \\
    \cite{duan2024pre} & Taukdial & MCI, HC &  77.5\% & Linguistic, acoustic, and data augmentation \\
\cite{fraser2019multilingual} & Pitt + Swedish dataset & MCI, HC & 63\% (Pitt)  & Multilingual word embeddings \\
    & & & 72\% (Swedish) & \\
\cite{jain2021exploring} & Pitt & AD, HC &  90.6\% & Domain-specific FastText word embeddings and 1D-CNN + BLSTM \\     
    \cite{kurtz2023early} & VAS & DM, MCI, HC & 74.7\% & Linguistic and acoustic features and ML classifiers \\
\cite{liu2022improving} & iFkyTek & AD, HC & 83.7\% & Transformer \\
\cite{orozco2024automatic} & Ivanova & AD, MCI, HC & 73.03\% & Data augmentation on audio files and CNN \\
\cite{sadeghian2021towards}& A custom conversational AD dataset&AD, HC&95.8\%& Linguistic, acoustic and ANN\\
    
\cite{Tamm2023Cross-Lingual}&ADReSS-M&AD, HC&82.6\%&Acouistic features and ANN\\
     \cite{wen2023revealing}
&A corpus from DementiaBank&AD, HC&92.2\%&PoS features and a transformer model\\

    \cite{ying2023multimodal} & NCMMSE & AD, MCI, HC & 89.1\% & Linguistic and acoustic features and SVM \\

    \bottomrule
    \hline
  \end{tabular}}
  \label{tab:related_work}
\end{table*}

\begin{itemize}
    \item \textbf{Pitt corpus}~\cite{becker1994dementia} is part of DementiaBank, a multimedia database designed to facilitate the study of individuals with dementia, supported by grants NIA AG03705 and AG05133. This dataset encompasses various types of conversational data, including tasks such as the Cookie Theft picture (CTP) description \cite{goodglass2001bdae}, fluency tasks, story recall, and sentence construction. For this study, data were exclusively drawn from Probable AD, PossibleAD, MCI, and HC participants for the CTP task.
    \item \textbf{Lu}~\cite{lanzi2023dementiabank} is an English-language dataset from DementiaBank, comprising conversations from 26 HC participants and 28 individuals with dementia, specifically focusing on CTP conversations from the United States.
    \item \textbf{VAS}~\cite{liang2022evaluating} consists of voice commands collected from 40 older adults, aged 65 or older, who were either HC participants or MCI participants. These participants were grouped based on their Montreal Cognitive Assessment (MoCA) \cite{hobson2015montreal} scores. The data was collected through Amazon Alexa, a Voice-Assistant System, and includes daily spontaneous voice commands issued by users to seek assistance with everyday tasks. 
    \item \textbf{Baycrest}~\cite{johnston2023spectral} includes retellings of the Cinderella story and other discourse tasks. For the Cinderella task, participants were provided with a storybook featuring Disney illustrations of the tale. The text was obscured, and pages deemed less relevant were taped together to prevent visibility.
    \item \textbf{Kempler}~\cite{kempler1987syntactic} originally was collected from individuals diagnosed with probable AD at the UCLA Geriatric Outpatient Clinic and the West Los Angeles Veterans Administration Hospital. This dataset contains  conversations from individuals with MCI, focusing on topics such as family, profession, personal history, and questions from a neuropsychological interview \cite{kempler1987syntactic}.
    \item \textbf{WLS} (The Wisconsin Longitudinal Study)~\cite{herd2014cohort} follows a random sample of Wisconsin high school graduates from 1957 (N = 10,317), born between 1938 and 1940. Initially focused on educational and occupational aspirations, later surveys in 1964, 1975, 1992, 2004, and 2011 increasingly focused on health and life course experiences as participants aged \cite{herd2014cohort}. 
    \item \textbf{Delaware}~\cite{lanzi2023dementiabank}, part of DementiaBank, investigates cognitive-linguistic features in neurotypical adults and individuals with MCI due to possible AD. It employs the DementiaBank discourse protocol, which includes picture description, story narrative, procedural discourse, and personal narrative tasks. 
    \item \textbf{Taukdial}~\cite{luz2024connected} consists of spontaneous speech samples from cognitively normal subjects and individuals with MCI, recorded while describing pictures. The dataset includes English and Chinese audio recordings.
    \item \textbf{Ivanova}~\cite{ivanova2022discriminating} comprises recordings from participants over 60 years old, all native European Spanish speakers with at least six years of primary education, ensuring literacy and minimizing cognitive load during a reading task. 
    \item \textbf{PerLA}~\cite{suarez2024alzheimer} is part of the Clinical Linguistics PERLACH Corpus. %
    Collected between 2012 and 2014, it includes 27 transcriptions from conversations with 21 AD patients, some interviewed twice.
    \item \textbf{Lu}~\cite{macwhinney2011aphasiabank}, available through DementiaBank, consists of interview recordings from 52 Mandarin-speaking AD patients. These recordings include tasks such as the CTP description, category fluency, and picture naming exercises \cite{qi2023noninvasive}.
    \item \textbf{NCMMSE}~\cite{ortiz2024deep}, from the National Conference on Man-Machine Speech Communication (NCMMSC), consists of audio recordings where participants engage in various tasks, such as picture descriptions and fluency exercises. It features 79 subjects with AD, 93 with MCI, and 108 HC. %
    \item \textbf{IFlytek}~\cite{liu2021spontaneous}, derived from the Predictive Challenge of AD in 2019, comprises transcripts of spontaneous speech collected during CTP description tasks, a common diagnostic tool for neurological disorders \cite{liu2021spontaneous}. Participants, ranging in age from 41 to 98, included 68 individuals with AD, 144 with MCI, and 111 healthy controls.
    \item \textbf{Dem@Care}~\cite{karakostas2016care} datasets consist of data collected through lab and home-based experiments conducted at the Greek Alzheimer’s Association for Dementia and Related Disorders in Thessaloniki, Greece, and participants' homes. These datasets include video and audio recordings alongside data from physiological sensors. Additionally, it incorporates data from sleep, motion, and plug sensors, offering a comprehensive view of participants' behavioral and physiological patterns. We used the Ds3, Ds5, and Ds7 darasets from Dem@Care project.
    \item \textbf{ADReSS-M}~\cite{luz2023multilingual} comprises spontaneous speech samples from cognitively normal individuals and AD patients. The training set includes English audio recordings of participants describing the CTP from the Boston Diagnostic Aphasia Examination. The test set features speech samples in Greek, where participants describe a different picture. For this study, we utilized only the test set including 24 HC, 22 AD, as the dataset creator advises against using the training set alongside other English datasets from DementiaBank due to potential overlap.
\end{itemize}

\subsection{Computational Approaches to Alzheimer's Detection}

Numerous studies have explored cognitive task-based approaches for Alzheimer’s detection, using speech and language analysis as key indicators of cognitive decline. These methods consider linguistic and paralinguistic features, such as lexical richness, syntactic complexity, fluency, and acoustic properties, to differentiate between AD, MCI, and HC across various languages and datasets \cite{shakeri2025natural}. We review related work focusing on single-language and multilingual conversation-based Alzheimer’s detection for both binary and multiclass classification, which are summarized in Table~\ref{tab:related_work}.

For binary AD detection, \citet{duan2024pre} analyze the multilingual Taukdial dataset, extracting language-agnostic speech features to classify MCI and HC. \citet{fraser2019multilingual} show that multilingual approaches outperform monolingual ones, achieving up to 72\% accuracy in MCI detection using narrative speech in English and Swedish. \citet{jain2021exploring} demonstrate that domain-specific word embeddings improve AD vs. HC classification on the Pitt corpus.

Regarding multiclass AD detection, \citet{orozco2024automatic} address the underrepresentation of Spanish speakers, using a CNN model on the Ivanova dataset from DementiaBank and achieving 73.03\% accuracy. Due to data limitations, they augmented the training set with synthetic samples. \citet{kurtz2023early} utilize the VAS dataset, achieving 74.7\% accuracy in differentiating DM, MCI, and HC, highlighting the potential of voice assistant systems for passive monitoring. \citet{ying2023multimodal} integrate acoustic and linguistic features from Wav2Vec2.0 and BERT for early AD detection in Chinese (NCMMSC dataset), achieving 89.1\% accuracy.

\citet{liu2022improving} propose a transformer-based model with a feature purification network, finding that semantic features generalize well across English and French. \citet{chen2023cross} show that paralinguistic features outperform pre-trained ones in cross-lingual AD detection, achieving 69.6\% accuracy when training on English and testing on Greek. \citet{Tamm2023Cross-Lingual} integrate acoustic features with demographic covariates, achieving 82.6\% accuracy for AD classification. \citet{wen2023revealing} identify 12 PoS features distinguishing AD from HC, reaching 92.2\% accuracy. Finally, \citet{sadeghian2021towards} propose a non-invasive, speech-based diagnostic tool for AD detection in clinical and home settings.

%% file: dementia-03.tex
\section{Dataset Creation}
\label{sec:dataset}

\begin{figure*}[t]
    \centering
    \includegraphics[width=0.8\textwidth]{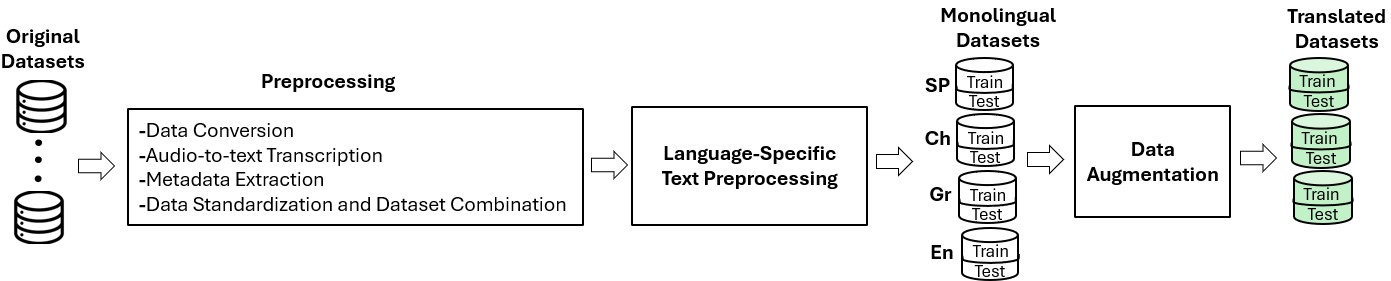} 
    \caption{Flowchart illustrating the process of dataset generation. }
    \label{fig:process}
\end{figure*}

\begin{table}[t]  
  \caption{Distribution of diagnostic categories (AD, MCI, HC) across languages with corresponding train and test splits. } %
  \captionshrink
  \centering
  \begin{tabular}{l|rrr|rrr}
    \toprule
    \multirow{2}{*}{\textbf{Language}} &  \multicolumn{3}{c|}{\textbf{Train Split}} & \multicolumn{3}{c}{\textbf{Test Split}} \\
    &  AD & MCI & HC & AD & MCI & HC \\
    \midrule
    Spanish &   78 & 70 & 156 &  20 & 18 & 39 \\
    Chinese &   93  & 200 & 187 &  23 & 51&47 \\
    Greek   &   266 & 51 & 87 & 66 & 13 & 22 \\
    English &  512 & 239 & 1450 & 63 & 60 & 87 \\
     \midrule
    \textbf{Total} &  \textbf{949} & \textbf{560} & \textbf{1880} & \textbf{172} & \textbf{142} & \textbf{195} \\
    \bottomrule
  \end{tabular}
  \label{tab:statistics_table}
\end{table}

\dataset{} is constructed from 16 individual datasets presented in Table~\ref{tab:datasetdescription}, which vary significantly in format, modality, and associated metadata. To create a unified multilingual dataset, we performed a series of preprocessing and normalization steps. This involved converting data to a standard format, transcribing audio data, extracting relevant metadata, standardizing data structures, and performing language-specific text cleaning. Figure~\ref{fig:process} illustrates the process. The overall goal was to create a consistent and well-structured dataset suitable for training and evaluating multilingual AD detection models.
A detailed breakdown of the number of conversation transcripts in each dataset, along with the train-test split statistics, is provided in Table~\ref{tab:statistics_table}.

\subsection{Preprocessing}

\subsubsection{Data Conversion}

The datasets exist in various formats, reflecting differences in modalities and data sources. All text-based datasets from DementiaBank\footnote{\url{https://dementia.talkbank.org/}} are provided in the CHAT format~\cite{macwhinney2017tools}. The Chinese iFlyTek text-based dataset is available in TSV format. All audio files are provided in WAV format.

\subsubsection{Audio-to-text Transcription}

Several datasets consisted solely of audio recordings. We transcribed them using a state-of-the-art Transformer-based multilingual speech recognition system.
Specifically, we employed the Whisper-Large multilingual model (\texttt{Large-V3})~\cite{radford2023robust} to transcribe recordings in Chinese, Greek, and Spanish. For speaker diarization, we implemented whisper-large-v3 along with NVIDIA's NeMo, as NeMo offers a robust framework for distinguishing speakers within a conversation. However, upon manual inspection, we found that the diarization results were often inaccurate, likely due to background noise or suboptimal audio quality. Consequently, we omitted the speaker diarization step and focused solely on transcribing the audio files. 

\subsubsection{Metadata Extraction}
We extracted all relevant information, including patient IDs, cognitive scores, demographic details, and conversational data—and standardized them into a unified format. Mini-Mental State Examination (MMSE) \cite{woodward2005mini} and MoCA \cite{nasreddine2005montreal} are two commonly reported cognitive scores in the datasets used for this study. MMSE is a widely used cognitive screening tool designed to assess cognitive impairment, particularly in dementia. A score below 24 typically indicates the need for further evaluation, though factors such as age, education, and cultural background can influence interpretation \cite{woodward2005mini}. MoCA is a cognitive screening tool designed to detect MCI by evaluating eight cognitive domains through rapid and sensitive tasks. A score below 26 indicates cognitive impairment, with the MoCA demonstrating higher sensitivity than the MMSE in detecting early cognitive decline \cite{nasreddine2005montreal}.
This information was typically provided in supplementary files, most commonly in Excel format. The dataset metadata, including details on the original data modality, underlying task, dataset name, language, test date, test duration, and testing environment, is also incorporated into the final normalized dataset. Comprehensive documentation of this process can be found in the GitHub repository accompanying this work.

\subsubsection{Data Standardization and Dataset Combination}

To facilitate integration into a single multilingual dataset, all extracted data and transcriptions were stored as JSONL files. Certain dataset-specific considerations were made during standardization: for the Pitt Corpus, only transcripts from the CTP task were included, and for the Chinese NCMMSE dataset, only the training set was used due to test set label unavailability. Datasets within the same language were combined, resulting in four standardized datasets in English, Spanish, Chinese, and Greek. To ensure fair comparisons, an 80\%-20\% train-test split was applied across all datasets, with the exception of the WLS dataset, which was used exclusively for training.

\paragraph{WLS dataset}
Given its extensive number of transcripts and its demonstrated utility in AD detection, as highlighted by \citet{guo2021crossing}, we incorporated the WLS dataset to enhance model performance even if it does not include explicit dementia diagnoses but provides cognitive test scores and health-related responses.

\begin{figure*}[t]
    \centering
    \includegraphics[width=0.8\textwidth]{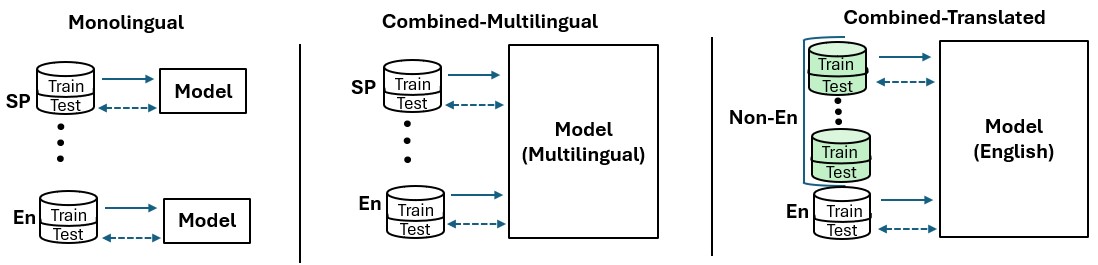} %
     \caption{Flowchart depicting the experimental scenarios. Monolingual;  Combined-Multilingual; Combined-Translated}
    \label{fig:variants}
\end{figure*}

Participants in the WLS dataset completed two verbal fluency tasks, naming words in the categories of animals and food within one minute. Verbal fluency tests are highly sensitive to AD detection, with a suggested screening cutoff of 15 words in the animal category \cite{guo2021crossing}. To establish a comparable control subgroup within WLS, we applied age- and education-adjusted verbal fluency thresholds: 16 words for participants under 60, 14 words for those aged 60–79, and 12 words for individuals over 79. Prior studies indicate that animal fluency scores below 15 are strongly indicative of AD, with a sensitivity of 0.88 and specificity of 0.96 \cite{guo2021crossing}. Since no normative data exist for the food category, we adopted the same thresholds as the animal category due to their similar score distributions, as suggested by \citet{guo2021crossing}.

\subsection{Language-Specific Text Preprocessing}

We developed four distinct text preprocessing pipelines, one for each language, to process the unified datasets generated during data collection. A major part of the preprocessing effort involved removing unnecessary symbols and marks in the transcripts, such as \texttt{*}, \&=laugh, and \&=nodes. Additionally, we created visualizations of transcript lengths and their frequency distributions to identify and remove potential outliers. The unified dataset comprises both the raw and preprocessed transcripts.

\subsection{Data Augmentation}

To broaden the usability of the dataset and facilitate addressing the research objectives outlined in the introduction, we augment it by including an English translation of Spanish, Greek, and Chinese texts.
We perform the translation with help of GPT-4 model accessed via the OpenAI API. Specifically, we used the prompt: 
\begin{small}
\begin{verbatim}
Translate the following {source_lang} text to {target_lang}: 
{text}
Translation:
\end{verbatim}
\end{small}
To ensure accuracy and consistency, we set the temperature parameter to zero, minimizing variability in responses.

\subsection{Dataset Variants}
\label{sec:dataset:variants}

We define three dataset variants to enable a range of evaluation scenarios, investigating the impact of multilingual data and translation on model performance. Figure~\ref{fig:variants} illustrates the corresponding experimental scenarios using these dataset variants.

\begin{itemize}
    \item \textbf{Monolingual:} Models are trained and evaluated solely on the train and test sets of the \emph{same} language. This represents a baseline scenario where only data from a single language is used.

    \item \textbf{Combined-Multilingual:}  Models are trained on a combination of the training sets from 
    \emph{all} languages.  This allows the model to potentially learn cross-lingual patterns.  Evaluation is performed on a single language's test set at each iteration.

    \item \textbf{Combined-Translated:} We utilize the machine-translated versions of all non-English datasets. The model is then trained on the combined English and English-translated training sets, while evaluation is performed on the translated test sets. This setting tests the effectiveness of using translation to create a unified English-only dataset.
\end{itemize}

%% file: dementia-04.tex
\section{Experiments}

This section details the experiments conducted on the unified MultiConAD  dataset. We explored two classification tasks: binary and multiclass classification. In the binary classification setting, the MCI group was excluded, concentrating on the distinction between HC and AD. The multiclass classification, which addresses a research gap identified in our recent literature review \cite{shakeri2025natural}, involved classifying instances into HC, MCI, and AD categories. To investigate multilingual patterns related to AD pathology, all classification models were trained on three different variants of the dataset---monolingual, combined-multilingual, and combined-translated---as described in Section~\ref{sec:dataset:variants}. %

\subsection{Research Questions}

We address the following research questions.

\begin{itemize}
    \item \textbf{RQ1}: How does the performance of Alzheimer's detection models differ when framed as a binary classification problem (AD vs. HC) compared to a multiclass problem (AD vs. MCI vs. HC)?

    \item \textbf{RQ2}: For multiclass classification, do the confusion matrices of the models reveal specific challenges in distinguishing between different stages of cognitive decline (AD vs. MCI)?

    \item \textbf{RQ3}: Does training language-specific models on a combined dataset in multiple languages (combined-multilingual) improve overall performance compared to training only on a single language (monolingual)?

    \item \textbf{RQ4}: How does the performance of models trained on translated and combined data (combined-translated) compare to models trained on the original language (monolingual) and on the combined dataset without translation (combined-multilingual)?

\end{itemize}

\begin{table*}[t]
  \centering
  \caption{Binary classification (AD vs. HC) results comparing Sparse and Dense text representations and classifiers (DT, RF, SVM, LR) across dataset variations. Best results are in boldface. The arrows denote a performance improvement \goodarrow~~ or degradation \badarrow~~ relative to the Monolingual setting.}
  \captionshrink
 \begin{tabular}{lcccccrccccrccccc}
    \toprule
    \multirow{2}{*}{\textbf{Language}} & \textbf{Text} & \multicolumn{4}{c}{\textbf{Monolingual}} && \multicolumn{4}{c}{\textbf{Combined-Multilingual}} && \multicolumn{4}{c}{\textbf{Combined-Translated}} \\
    \cline{3-6} \cline{8-11} \cline{13-16}
          &  \textbf{Repr.} 
          & DT    & RF    & SVM   & LR & 
          & DT    & RF    & SVM   & LR &
          & DT    & RF    & SVM   & LR \\
    \midrule
    \multirow{2}{*}{Spanish} & Sparse 
        & 0.73  & 0.73  & \textbf{0.78}  & 0.78 & 
        & 0.80 \goodarrow & 0.76 \goodarrow& 0.66 \badarrow & \textbf{0.80}\goodarrow & 
        & 0.75\goodarrow&\textbf{0.80}\goodarrow &0.75 \badarrow &0.73 \badarrow \\
      & Dense 
      & 0.71  & 0.71  & \textbf{0.80} & 0.80 & 
      & 0.66 \badarrow & 0.78 \goodarrow& 0.66 \badarrow & \textbf{0.80} \nochange&
      & 0.59 \badarrow& 0.76\goodarrow&\textbf{0.78}\badarrow &0.76\badarrow  \\
    \midrule
    \multirow{2}{*}{Chinese} & Sparse 
    & 0.67  & 0.69  & \textbf{0.70}  & 0.70 & 
    & 0.67\nochange  & \textbf{0.69}\nochange & 0.67 \badarrow & \textbf{0.69}\badarrow& 
    & 0.70\goodarrow &\textbf{0.90} \goodarrow&0.89 \goodarrow&0.84  \goodarrow \\
   & Dense 
   & 0.69  & 0.76  & \textbf{0.83} & 0.80  & 
   & 0.76 \goodarrow &\textbf{ 0.86} \goodarrow & 0.67\badarrow & 0.81 \goodarrow&
   & 0.66 \badarrow&\textbf{0.84}\goodarrow &0.80 \badarrow&0.84 \goodarrow \\
    \midrule
    \multirow{2}{*}{Greek} & Sparse 
    & 0.68  & \textbf{0.78} & 0.77  & 0.78 & 
    & 0.68 \nochange & \textbf{0.76} \badarrow& 0.60 \badarrow & 0.73 \badarrow&
    & 0.58 \badarrow&0.67 \badarrow &\textbf{0.69} \badarrow& 0.51\badarrow \\
    & Dense 
    & 0.65  & 0.75  & \textbf{0.78} & 0.77 & 
    & 0.64  \badarrow& \textbf{0.75} \nochange & 0.75\badarrow & 0.73 \badarrow& 
    & 0.62\badarrow&0.66\badarrow  &\textbf{0.70} \badarrow &0.64 \badarrow  \\
    \midrule
    \multirow{2}{*}{English} & Sparse 
    & 0.73  & \textbf{0.78}        & 0.77 & 0.75  & 
    & 0.67 \badarrow & 0.74 \badarrow & 0.58&\textbf{0.75}&
    & 0.73 \nochange&0.74 \badarrow &\textbf{0.76} &0.61 \badarrow\\
     & Dense
     & 0.65  & 0.75  & \textbf{0.81} & 0.79  & 
     & 0.67 \goodarrow & \textbf{0.77}\goodarrow  & 0.58 \badarrow& 0.67 \badarrow &
     & 0.71\goodarrow& 0.73\badarrow &\textbf{0.83}\goodarrow &0.70 \badarrow\\
    \bottomrule
    \end{tabular}%
  \label{tab:Binary-class_Classification_Result}%
\end{table*}%

\begin{table*}[t]
  \centering  
  \caption{Multiclass classification (AD vs. MCI vs. HC) results comparing Sparse and Dense text representations and classifiers (DT, RF, SVM, LR) across dataset variations. Best results are in boldface. The arrows denote a performance improvement \goodarrow~~ or degradation \badarrow~~ relative to the Monolingual setting.}
  \captionshrink
    \begin{tabular}{lcccccrccccrccccc}
    \toprule
    \multirow{2}{*}{\textbf{Language}} & \textbf{Text} & \multicolumn{4}{c}{\textbf{Monolingual}} && \multicolumn{4}{c}{\textbf{Combined-Multilingual}} && \multicolumn{4}{c}{\textbf{Combined-Translated}} \\
    \cline{3-6} \cline{8-11} \cline{13-16}
          &  \textbf{Repr.} 
          & DT    & RF    & SVM   & LR & 
          & DT    & RF    & SVM   & LR &
          & DT    & RF    & SVM   & LR \\
    \midrule
    \multirow{2}{*}{Spanish} & Sparse 
        & 0.61 & 0.60 & \textbf{0.61} & 0.61 &
        & 0.51 \badarrow & \textbf{0.62} \goodarrow & 0.51 \badarrow & 0.58 \badarrow &
        & 0.47 \badarrow & \textbf{0.58} \badarrow & 0.56 \badarrow & 0.56 \badarrow \\
    & Dense 
        & 0.52 & 0.61  & \textbf{0.61} & 0.61 & 
        & 0.47 \badarrow & \textbf{0.61} \nochange & 0.61 \nochange & 0.57 \badarrow &
        & \textbf{0.60} \goodarrow & 0.58 \badarrow & 0.56 \badarrow & 0.51\badarrow  \\
    \midrule
    \multirow{2}{*}{Chinese} & Sparse 
        & 0.36 & 0.35 & \textbf{0.40} & 0.39 &
        & \textbf{0.42} \goodarrow & 0.39 \goodarrow &0.39 \badarrow & 0.40 \goodarrow & 
        & 0.45 \goodarrow & 0.59 \goodarrow & \textbf{0.68} \goodarrow & 0.62 \goodarrow\\
    & Dense 
        & 0.51 & 0.58  & \textbf{0.59} & 0.56 &
        & 0.43 \badarrow & \textbf{0.62}\goodarrow & 0.60  \goodarrow& 0.60 \goodarrow& 
        & 0.43 \badarrow & \textbf{0.64}\goodarrow & 0.60 \goodarrow& 0.45 \badarrow \\
    \midrule
    \multirow{2}{*}{Greek} & Sparse 
        & 0.59 & \textbf{0.74} & 0.67 & 0.71 &
        & 0.57 \badarrow & \textbf{0.71} \badarrow& 0.53 \badarrow& 0.66 \badarrow&
        & 0.64 \goodarrow & 0.65 \badarrow & \textbf{0.69}\goodarrow & 0.60 \badarrow \\
    & Dense 
        & 0.54 & 0.66  & \textbf{0.73} & 0.73 &
        & 0.54  \nochange & 0.66 \nochange& 0.65 \badarrow & \textbf{0.67} \badarrow & 
        & \textbf{0.62} \goodarrow & 0.61 \badarrow & 0.60 \badarrow & 0.42 \badarrow  \\
    \midrule
    \multirow{2}{*}{English} & Sparse 
        & 0.59 & 0.62 & \textbf{0.65} & 0.65 &
        & 0.59 \nochange & 0.58 \badarrow& 0.41 \badarrow& \textbf{0.66} \goodarrow&
        & 0.50 \badarrow & 0.61 \badarrow& \textbf{0.66}\goodarrow & 0.64 \badarrow \\
    & Dense 
        & 0.51 & 0.62 & \textbf{0.65} & 0.63 &
        & 0.50 \badarrow & 0.62 \nochange & \textbf{0.65} \nochange & 0.63 \nochange& 
        & 0.50 \badarrow & 0.57 \badarrow  & \textbf{0.66} \goodarrow& 0.41 \badarrow \\
    \bottomrule
    \end{tabular}%
  \label{tab:multiclass_Classification_Result}%
\end{table*}%

\subsection{Baseline methods}
\label{sec:baselines}

We consider two text representations, sparse and dense, and a diverse set of established classification algorithms.
We chose a TF-IDF weighting for sparse representation due to its effectiveness in highlighting less common, more informative words. This approach has been utilized in similar studies, such as \cite{adhikari2022exploiting,searle2020comparing,martinc2020tackling,Shakeri2024ICAPAI}. We conducted experiments on all datasets both with and without stopword removal. However, we observed that eliminating stopwords led to a decline in model performance. Based on this finding, we decided to forgo stopword removal in our final approach. For dense representation, we employ a multilingual embedding model multilingual \texttt{intfloat/multilingual-e5-large} \cite{wang2024multilingual}, which is initialized from xlm-roberta-large and trained on a mixture of multilingual datasets. It supports 100 languages from xlm-roberta. The e5-large model, based on XLM-RoBERTa, features 24 layers, 16 attention heads, and a 1024 hidden size. It uses GELU activation, 0.1 dropout, and a 250,002-token vocabulary, with a 4096 intermediate size for capturing complex linguistic structures.

After obtaining feature representations through the aforementioned sparse and dense approaches, the resulting features were fed into four machine learning classifiers: Decision Tree (DT), Random Forest (RF), Support Vector Machine (SVM), and Logistic Regression (LR). To optimize the performance of classifiers, we employed grid search in combination with 5-fold cross-validation to evaluate various hyperparameter configurations. DT was tuned for different depths (max\_depth: 10, 20, 30), while the RF was evaluated with varying numbers of estimators (n\_estimators: 50, 100, 200). For SVM, we experimented with different regularization strengths (C: 0.1, 1, 10) and kernel types (linear, rbf). Similarly, LR was tested with different C values (0.1, 1, 10) to control regularization. The best hyperparameters were chosen based on accuracy, and the final model was trained on the full training set. %

\subsection{Results}

\begin{figure*}[t]
    \centering    
    \begin{subfigure}[b]{0.24\textwidth}
        \centering
        \includegraphics[width=\textwidth]{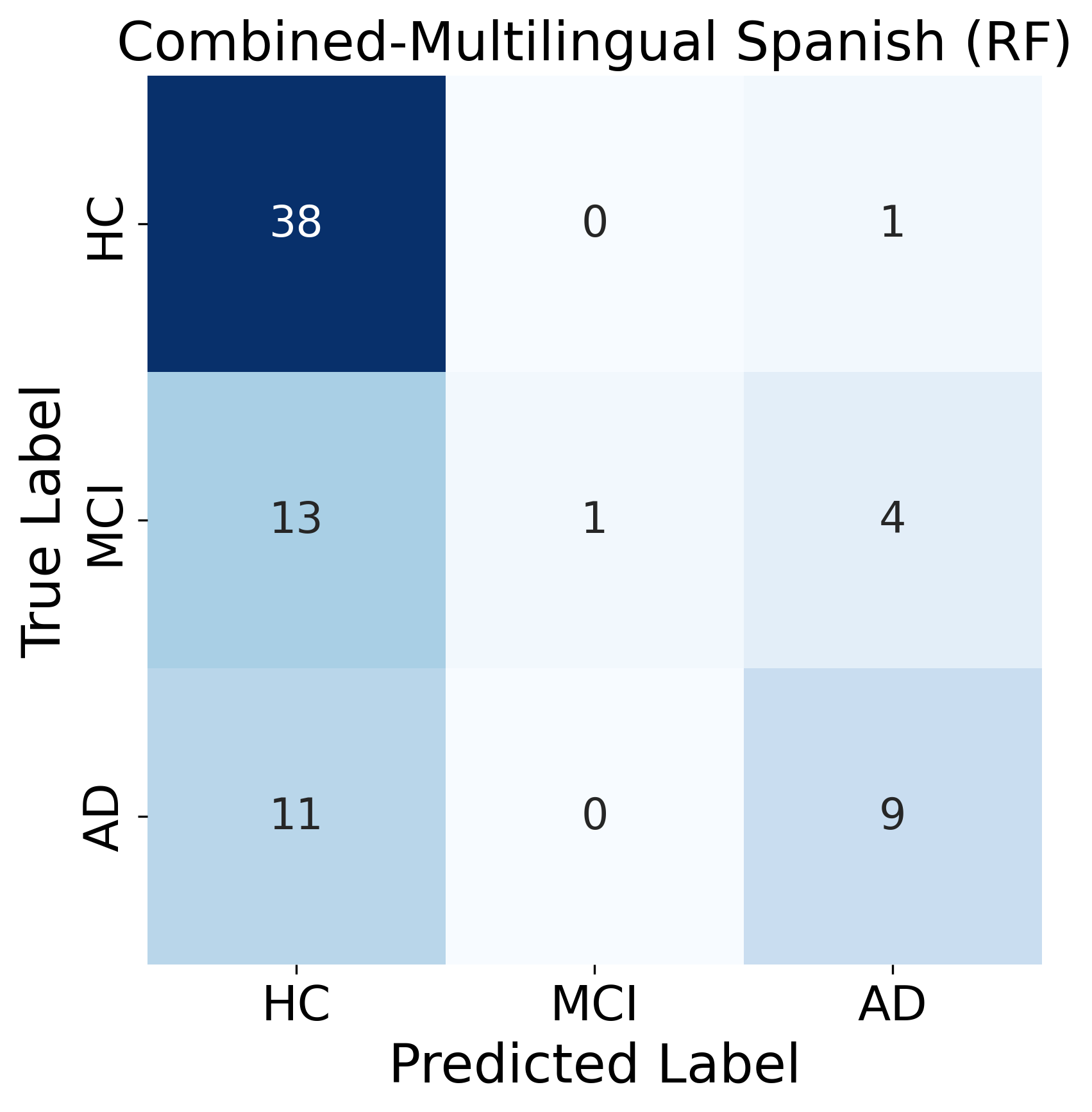}
        \caption{}
    \end{subfigure}
    \hfill
    \begin{subfigure}[b]{0.24\textwidth}
        \centering
        \includegraphics[width=\textwidth]{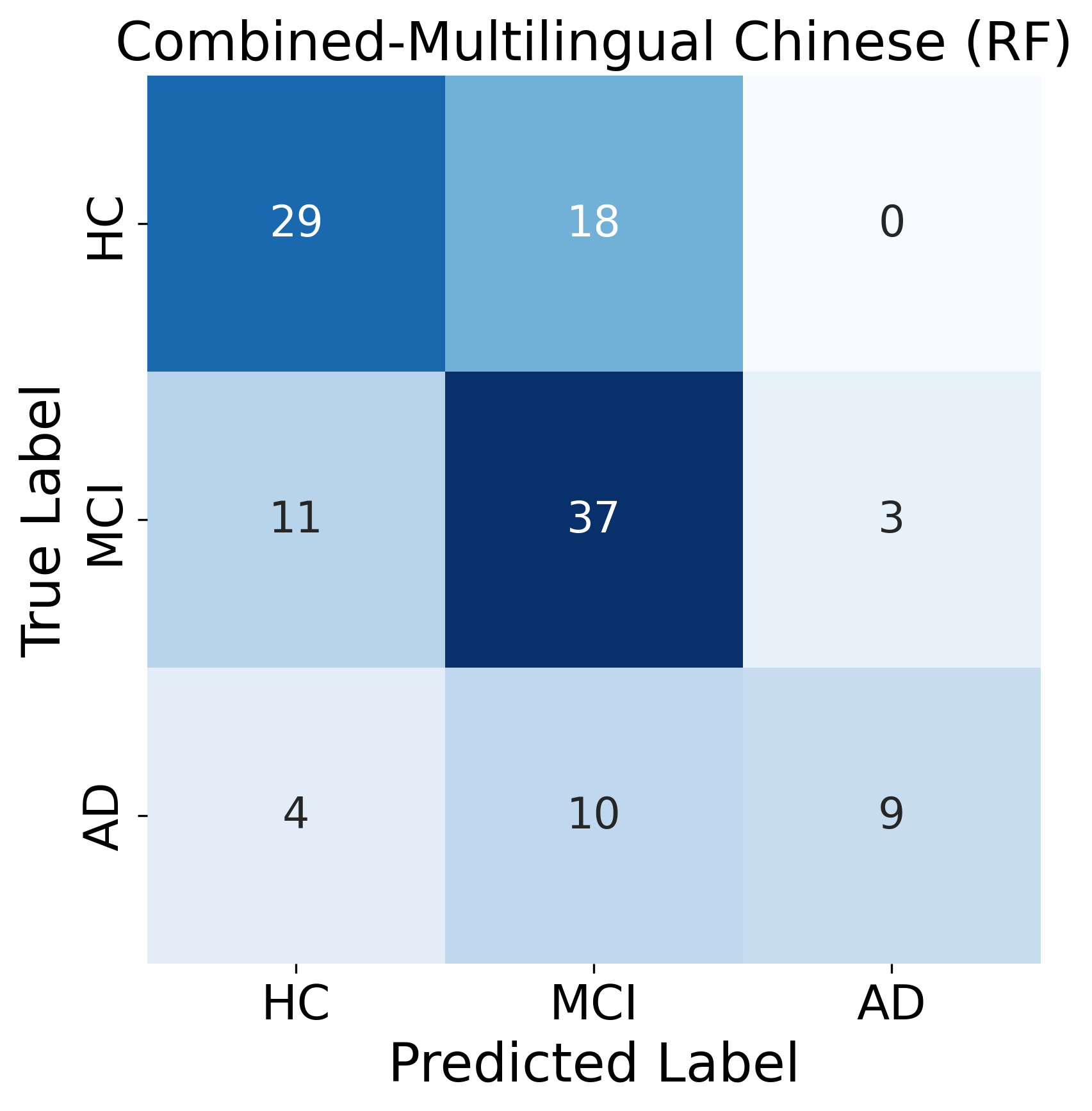}
        \caption{}
    \end{subfigure}
    \hfill
    \begin{subfigure}[b]{0.24\textwidth}
        \centering
        \includegraphics[width=\textwidth]{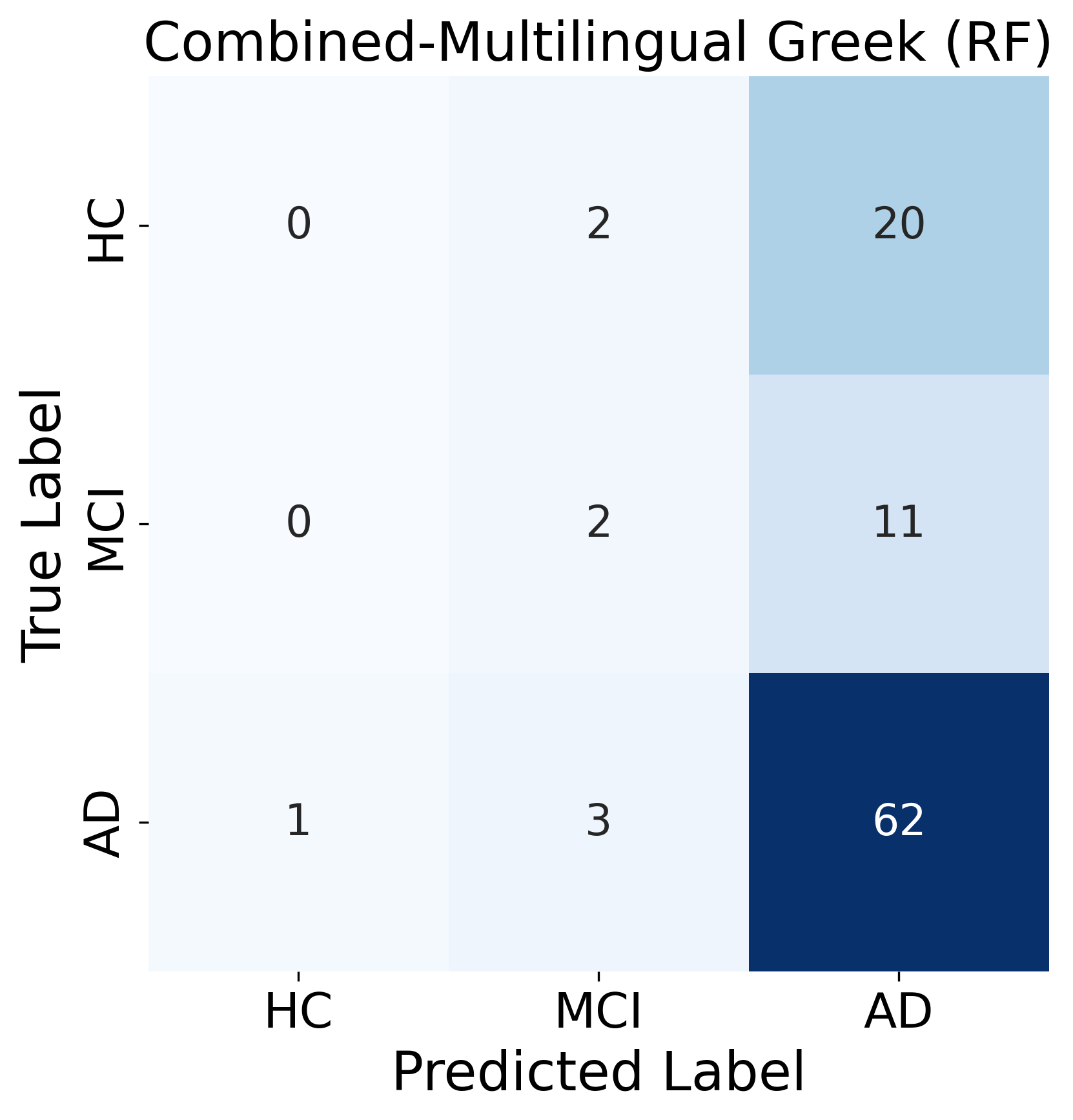}
        \caption{}
    \end{subfigure}
\begin{subfigure}[b]{0.24\textwidth}
        \centering
        \includegraphics[width=\textwidth]{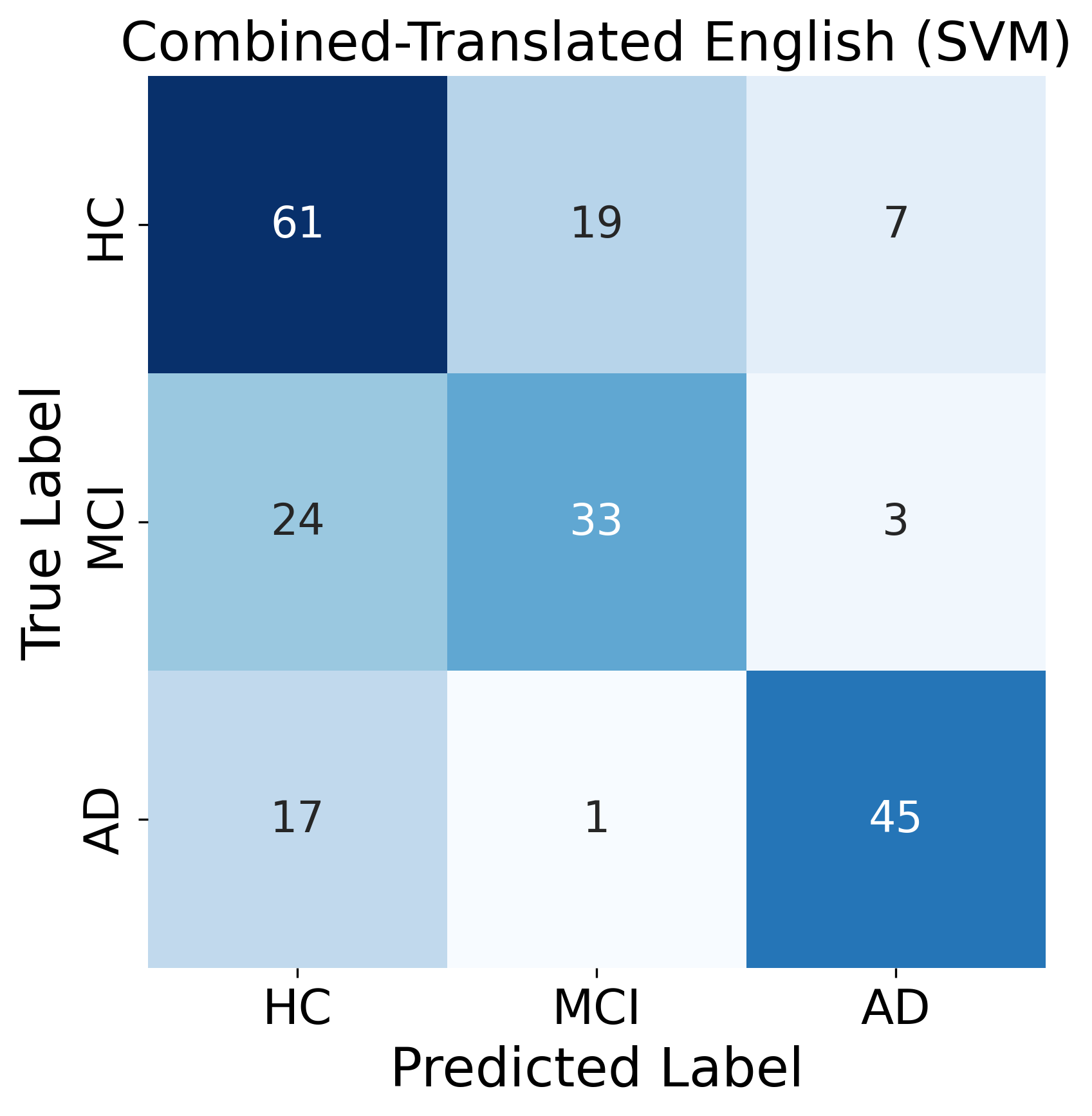}
        \caption{}
    \end{subfigure}
    \hfill
    \caption{Confusion matrices for multiclass classification using the best-performing dataset variant and classifier for each language.}
      \label{confusion matrices}
\end{figure*}

\subsubsection{Binary vs. Multiclass Classifiation}

We first ask how the performance of the models compare when Alzheimer's detection is framed as a binary vs. multiclass classification problem (RQ1). Tables \ref{tab:Binary-class_Classification_Result} 
 and \ref{tab:multiclass_Classification_Result} present the results of binary and multiclass classification based on both dense and sparse representations, evaluated across the monolingual, multilingual-combined, and combined-translated settings. The experimental results indicate that AD detection is a significantly easier problem when formulated as a binary classification task (AD vs. HC) compared to a multiclass classification (AD vs. MCI vs. HC). Across all models and datasets, binary classification consistently yields higher accuracy, with peak performance reaching 0.90 on the combined-translated dataset (Chinese, sparse, RF). In contrast, multiclass classification demonstrates a decline in performance, with the highest observed accuracy not exceeding 0.74 on the monolingual dataset (Greek, sparse, RF). These findings suggest that distinguishing between AD and HC is relatively straightforward, whereas differentiating between AD, MCI, and HC presents greater challenges. The reduced performance in the multiclass setting is likely attributable to the clinical and cognitive overlap between MCI and both AD and HC, leading to increased misclassification rates. This is reflected in the consistently lower accuracy observed in multiclass classification compared to the binary classification scenario.

In classification tasks, predictions made by a machine-learned classifier are subject to error, which manifests as two primary types: Type I error (false positives) and Type II error (false negatives). In AD detection, a Type I error means incorrectly classifying a HC individual as having AD, while a Type II error means incorrectly classifying an individual with AD as healthy. While both types of error are undesirable, the consequences of a Type II error in a clinical setting are far more severe. Missing a diagnosis of AD (a Type II error) delays potential treatment and intervention, potentially leading to worse patient outcomes. Conversely, a Type I error, while still causing concern and requiring further testing, allows for proactive monitoring and reduces the risk of undetected disease progression. Therefore, while we acknowledge both error types, this study prioritizes Type II error, which we measure in terms of \emph{false negative rate}:
the proportion of non-HC participants incorrectly classified as HC, relative to the total number of HC participants (i.e., FN/(FN+TP)). Table \ref{tab:Mis_Classification_Result} presents the false negative rates for the best-performing classifier across all languages using the dense representation model for multiclass classification. The results for monolingual and combined-translated models are based on SVM, while the Multilingual-Combined model results are based on the RF classifier. In multiclass classification, Spanish has the highest error (0.78), followed by English (0.37), while Greek (0.06) and Chinese (0.24) perform better. The combined-multilingual dataset reduces errors, especially in Spanish (0.63) and Greek (0.01), but translation increases errors in some cases. Overall, false negative rate is high in the multiclass setting, particularly due to the difficulty of distinguishing MCI from HC, while multilingual training provides inconsistent improvements across languages.

\begin{table}[t]
  \centering  
  \caption{False negative rate in multiclass classification using SVM for Monolingual and Combined-Translated, and RF for Combined-Multilingual. The arrows denote a performance improvement \goodarrow~~ or degradation \badarrow~~ relative to the Monolingual setting.}
  \captionshrink
  \begin{tabular}{lccc}
    \toprule
    \multirow{2}{*}{\textbf{Language}}  & \multirow{2}{*}{\textbf{Monolingual}} & \textbf{Combined-} & \textbf{Combined-} \\ 
    & & \textbf{Multilingual} & \textbf{Translated} \\
    \midrule
    Spanish  & 0.78 &  0.63 \goodarrow & 0.71 \goodarrow \\
    Chinese &  0.24 &  0.20 \goodarrow & 0.31 \badarrow  \\
    Greek  &   0.06 &  0.01 \goodarrow & 0.13 \badarrow  \\
    English  &    0.37 &  0.48 \badarrow &  0.33 \goodarrow \\
    \bottomrule
  \end{tabular}
  \label{tab:Mis_Classification_Result}%
\end{table}

\subsubsection{Different Stages of Cognitive Decline}

Next, we take a closer look at the difficulty of in differentiating between various stages of cognitive decline in the multiclass setting.
Figure~\ref{confusion matrices} presents the confusion matrices for the four languages, using the models with the lowest false negative rate for each language, as shown in Table~\ref{tab:Mis_Classification_Result}. Specifically, Spanish, Chinese, and Greek are based on combined-multilingual with RF, while English is based on Translated-Combined with SVM. A common issue across all models is the misclassification of MCI, which is frequently confused with either HC or AD, highlighting the progressive and overlapping nature of cognitive decline. In the Spanish model, HC cases are classified with high accuracy, but distinguishing MCI and AD remains problematic. Similarly, in the Chinese model, a substantial number of HC cases are misclassified as MCI, and while MCI is relatively well classified, some cases are incorrectly labeled as AD. The Greek model exhibits the most significant challenge in HC classification, with HC cases frequently misclassified as AD, suggesting potential dataset imbalances or language-specific difficulties in speech-based diagnosis. In contrast, AD cases in Greek are classified with high accuracy, indicating clearer distinguishing features at later disease stages. The English model, trained on the combined-translated dataset, demonstrates strong performance in identifying HC cases but struggles to distinguish MCI, with a considerable number of MCI cases being misclassified as HC or AD. Overall, the results underscore the inherent difficulty in differentiating MCI from both HC and AD, with varying degrees of classification accuracy across languages.

Another key observation is that the Spanish model predominantly predicts HC across all true labels, indicating a strong bias that limits its ability to distinguish MCI and AD cases. In contrast, the Greek model exhibits severe misclassification toward AD, suggesting overconfidence in AD predictions while misclassifying HC and MCI cases. This pattern may result from class imbalance or insufficient training data, which could be influencing the model’s decision-making process.

\subsubsection{Monolingual vs. Multilingual Dataset}
Tables \ref{tab:Binary-class_Classification_Result} and \ref{tab:multiclass_Classification_Result} highlight the results from the best-performing classifier in bold for each language and dataset variant. 
In binary classification, the impact of multilingual training varies across languages, with some benefiting while others experience a decline. The best-performing model for Chinese shows a notable improvement when trained on the combined-multilingual dataset, increasing accuracy from 0.83 to 0.86. However, Spanish sees no change, maintaining a peak accuracy of 0.80 in both monolingual and multilingual settings. In contrast, Greek and English experience declines, with Greek dropping from 0.78 to 0.76 and English decreasing from 0.81 to 0.77. These results indicate that multilingual training does not consistently enhance binary classification performance and that its effectiveness is highly language dependent.

For multiclass classification, training language-specific models on the combined-multilingual dataset generally leads to performance improvements, except for Greek. Spanish achieves a slight increase in accuracy, improving from 0.61 in the monolingual setting to 0.62 in the multilingual setting. Similarly, Chinese benefits from a boost from 0.59 to 0.62, while English sees a marginal gain from 0.65 to 0.66. However, Greek, which performs best in a monolingual setting with an accuracy of 0.74, drops to 0.71 when trained in a multilingual setting. This suggests that while multilingual training is beneficial for certain languages in multiclass classification, it does not universally enhance performance across all cases.\\

\subsubsection{Translated Dataset}

In binary classification, training on the combined-translated dataset has varying effects across languages, occasionally boosting performance but often leading to declines. The best-performing model for English benefits from translation-based augmentation, increasing accuracy to 0.83 compared to 0.81 in the monolingual setting and 0.77 in the combined-multilingual setting. Chinese experiences a raise, increasing from 0.86 with combined-multilingual training and 0.83 in monolingual to 0.90 with combined-translation. Greek sees the most significant decline, falling from 0.78 in the monolingual setting and 0.76 in the combined-multilingual setting to just 0.70 with translation-combined. Spanish remains stable at 0.80 across all datasets, indicating that translation does not provide additional value for this language.

In multiclass classification, training on the combined-translated dataset enhances accuracy for Chinese, increasing to 0.68 compared to 0.59 in the monolingual setting and 0.62 in the combined-multilingual setting. English maintains its performance at 0.66, matching the combined-multilingual setting and slightly improving over the 0.65 accuracy in the monolingual setting. However, Spanish sees a slight decline (0.60 vs. 0.61 monolingual, 0.62 combined-multilingual), while Greek experiences a more noticeable drop (0.69 vs. 0.74 monolingual, 0.71 combined-multilingual), suggesting that translation does not provide uniform benefits across languages.

%% file: dementia-05.tex
\balance
\section{Conclusion AND FUTURE WORK}
\label{sec:concl}

This paper presents a geographically diverse, multilingual conversational dataset for Alzheimer's detection, comprising 16 publicly available datasets, featuring a range of cognitive assessment tasks, in English, Spanish, Chinese, and Greek. Additionally, an English-translated version of the multilingual dataset has been created to evaluate the impact of translation on Alzheimer's detection across the included languages. The main contributions include unifying datasets with varying formats into a consistent text format. The work also investigates how model framing (binary vs. multiclass) and dataset composition (monolingual vs. combined-multilingual, vs. combined-translated) affect Alzheimer’s detection performance and model differentiation between cognitive decline stages.

The results of this study highlight the fundamental challenges in Alzheimer’s detection when transitioning from a binary to a multiclass classification approach. While distinguishing between
cases of Alzheimer’s Disease (AD) and Healthy Controls (HC) is often achievable with reasonable accuracy, achieving the same level of accuracy in a multiclass setting that also includes patients with Mild Cognitive Impairment (MCI) proves significantly more difficult. This is largely due to the clinical and cognitive overlap between MCI and the other two categories, leading to increased misclassification rates. Our findings also shed light on the potential benefits and limitations of multilingual and translated datasets. While some languages, such as Chinese, show notable improvements with combined-multilingual or translated training, others, such as Greek, experience a decline in performance. This suggests that the effectiveness of multilingual learning is highly language-dependent, likely influenced by underlying cognitive tasks conducted in the dataset collection step, dataset size, linguistic features, and model adaptation. Furthermore, while translation can enhance performance in certain cases, it does not provide a universal advantage across languages, sometimes introducing additional classification errors, e.g., in Greek. These findings emphasize the need for tailored approaches in multilingual Alzheimer’s detection, considering language-specific challenges and dataset characteristics to optimize diagnostic accuracy.

This multilingual conversational dataset opens up significant avenues for future research in Alzheimer's detection. While our results demonstrate the feasibility of automatic detection, the substantial headroom for improvement suggests several promising directions.
First, future research may focus on optimizing language-specific approaches, tailoring methods to the unique characteristics of each language. A deeper investigation into syntactic, semantic, and lexical features could reveal subtle, language-specific indicators of cognitive decline. Second, advanced preprocessing techniques to explicitly identify participant roles (i.e., patient vs. interviewer) within the conversations could help create more nuanced models that specifically target the patient's language use.
Third, investigating cross-lingual patterns and leveraging transfer learning techniques could enhance model generalization and robustness, particularly for languages with limited data. Finally, exploring advanced data augmentation methods to address the class imbalance across diagnostic groups (HC, MCI, and AD) could help improve classification performance.